Short Paper*

# Quadcopter Position Hold Function using Optical Flow in a Smartphone-based Flight Computer

Noel P. Caliston
Iloilo State University of Fisheries Science and Technology, Philippines
npcaliston@gmail.com
(corresponding author)

Chris Jordan G. Aliac
College of Computer Studies, Cebu Institute of Technology University, Philippines
chris.aliac@cit.edu

James Arnold E. Nogra
College of Computer Studies, Cebu Institute of Technology University, Philippines
james.nogra@cit.edu

## Abstract

*Purpose* – This paper explores the capability of smartphones as computing devices for a quadcopter, specifically in terms of the ability of drones to maintain a position known as the position hold function. Image processing can be performed with the phone's sensors and powerful built-in camera.

*Method* – Using Shi-Tomasi corner detection and the Lucas-Kanade sparse optical flow algorithms, ground features are recognized and tracked using the downward-facing camera. The position is maintained by computing quadcopter displacement from the center of the image using Euclidian distance, and the corresponding pitch and roll estimate is calculated using the PID controller.

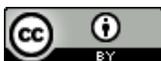

*Results* – Actual flights show a double standard deviation of 18.66 cm from the center for outdoor tests. With a quadcopter size of 58cm x 58cm used, it implies that 95% of the time, the quadcopter is within a diameter of 96 cm. For indoor tests, a double standard deviation of 10.55 cm means that 95% of the time, the quadcopter is within a diameter of 79 cm.

*Conclusion* – Smartphone sensors and cameras can be used to perform optical flow position hold functions, proving their potential as computing devices for drones.

*Recommendations* – To further improve the positioning system of the phone-based quadcopter system, it is suggested that potential sensor fusion be explored with the phone's GNSS sensor, which gives absolute positioning information for outdoor applications.

*Research Implications* – As different devices and gadgets are integrated into the smartphone, this paper presents an opportunity for phone manufacturers and researchers to explore the potential of smartphones for a drone use-case.

## INTRODUCTION

A quadcopter, multi-copter, or multirotor is a type of drone or UAV (Unmanned Aerial Vehicle) that has gained attention in the research community in the past decade. The rapid development of drones has paved the way for them to become a part of our society. It found many different applications in leisure, military, academic, industrial, and commercial (Rovira-Sugranes et al., 2022; Mohsan et al., 2023). Quadcopters can either be autonomous or controlled via a radio transmitter and receiver. These flying machines include many advanced features in their repertoire, such as altitude hold, position hold, return to home, and waypoint navigation. These features can usually be found in expensive commercial drones. However, academics developed custom drones for custom purposes, as commercial drones are difficult to modify to fit a specific purpose, aside from being too expensive.

Custom-built quadcopters are usually used for research or leisure. They are equipped with flight controllers and flight computers that determine the attitude of the flying machine. Many commercial flight controllers are available for custom drones, while flight computers are usually made of Raspberry Pi boards programmed to give commands to the flight controller (Shankar et al., 2022).

So far, most applications of mobile phones on drones have been limited to remotely controlling them or for monitoring purposes (Russell, 2016; Benhadhria et al., 2021; Shah et al., 2023). In the past, attempts to create an Android-based mobile phone flight controller only involved controlling drone attitudes such as roll, pitch, and yaw (Bergkvist, 2013). They need advanced features such as altitude or position hold. Mobile phones are equipped with devices and sensors such as a gyroscope, accelerometer, magnetometer, barometer, GPS, high-quality cameras, Wi-Fi, Bluetooth, and a powerful microprocessor to create advanced drone features. Thus, it is a cheaper substitute for combined commercial flight controllers, Raspberry Pi boards, and other sensors.

Position hold is the ability of a quadcopter to maintain position. This functionality is usually accomplished using a GPS module for outdoor or an optical flow sensor for indoor navigation. This paper demonstrates how a low-price mobile phone-based flight computer can perform the position-hold function of a quadcopter using a mobile phone camera. Image processing and attitude computations are executed on the phone. The processing power of this device opens the potential of a mobile phone as a UAV flight controller.

In the present study, a custom-built quadcopter was developed for farm crops and weed monitoring. This quadcopter was designed to be autonomous, using an Arduino-based MCU and an Android smartphone as its flight controller. The main idea is to utilize built-in mobile phone sensors to perform flight maneuvers. Trial and error testing results show low-priced mobile phones' GNSS updates to only 1Hz and are noisier than U-Blox receivers (Zangenehnejad & Gao, 2021). Thus, the phone's camera is a potential supplement for the smartphone-based position-hold function. The idea of including an optical flow position-hold feature is envisioned to compensate for the low update rate of GNSS in consumer devices. These techniques allow a potential sensor fusion between the mobile phone's GNSS sensor and the optical flow using its camera.

## METHODOLOGY

Android phones were used during the execution of this paper. Figure 1 shows a single iteration of the optical flow-based position-hold function using the Android phone camera. The algorithm is divided into several stages, from image capture until the required pitch and roll commands are computed and fed to the quadcopter for execution. Image processing computations such as corner detection and motion tracking are performed using OpenCV (Bradski, 2000).

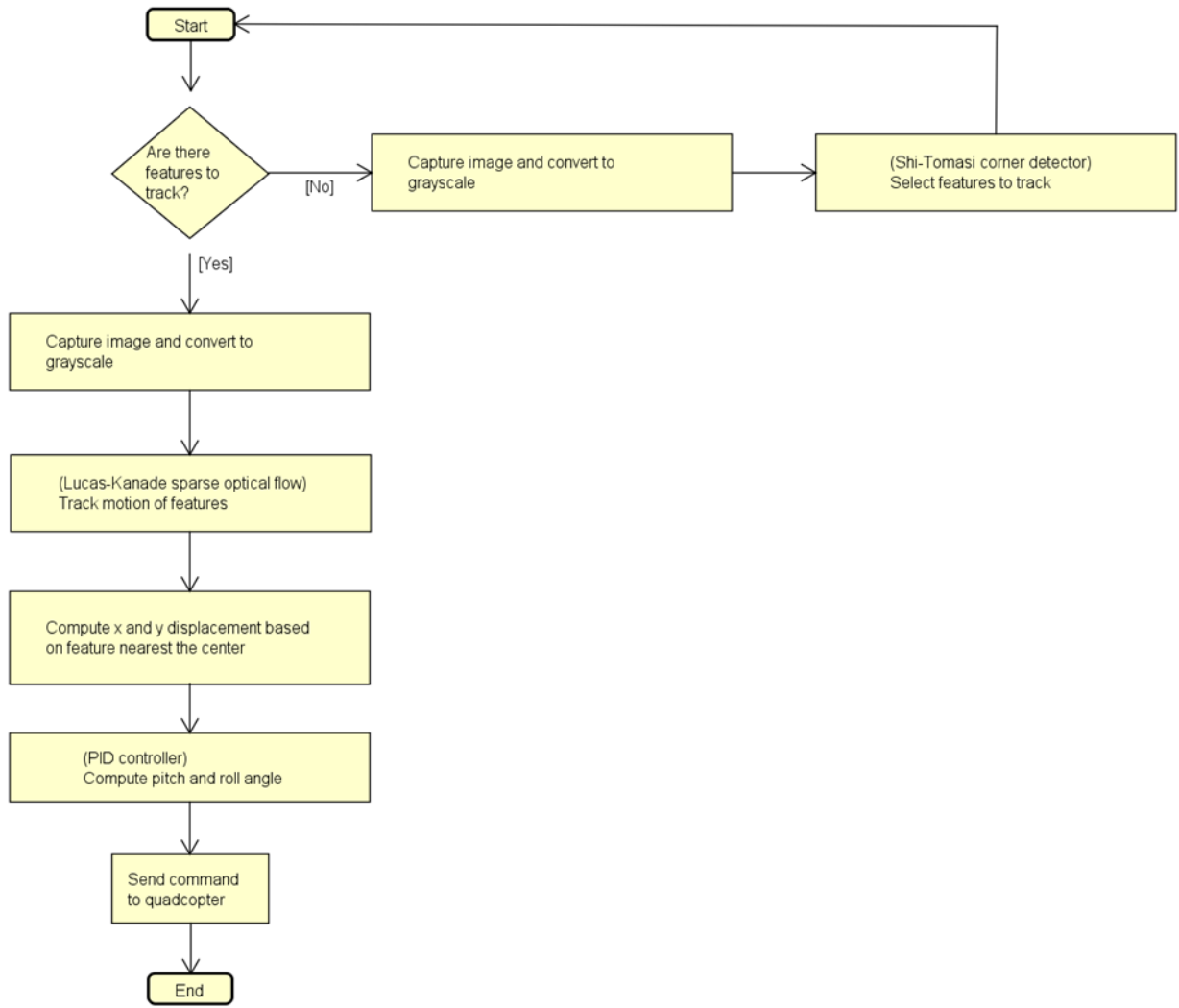

*Figure 1.* Quadcopter optical flow position-hold flowchart.

### Feature detection and motion tracking

The optical flow position-hold feature of this quadcopter starts by reading frames of images from the mobile phone camera. An initial frame is converted into a grayscale image. Each image is 640x480 in resolution. From this image, features are selected based on the Shi-Tomasi corner detector scoring function (Shi & Tomasi, 1994; Zhang et al., 2023):

$$R = min\ (\lambda 1,\ \lambda 2) \qquad Equation\ 1$$

Features in images are points of interest that present rich image content information. Depending on R in (1), a window can be classified as a flat, edge, or corner. If R is greater than the threshold, then it is a corner. The image fed into the corner detector is constrained only to include corners detected in the center part of the image. The captured image is divided into four parts, vertically and horizontally, considering only the center portion, ensuring the initial features for position-hold are located near the center, as shown in Figure 2a. This allows the quadcopter not to drift by making the drone follow corners at the image's sides.

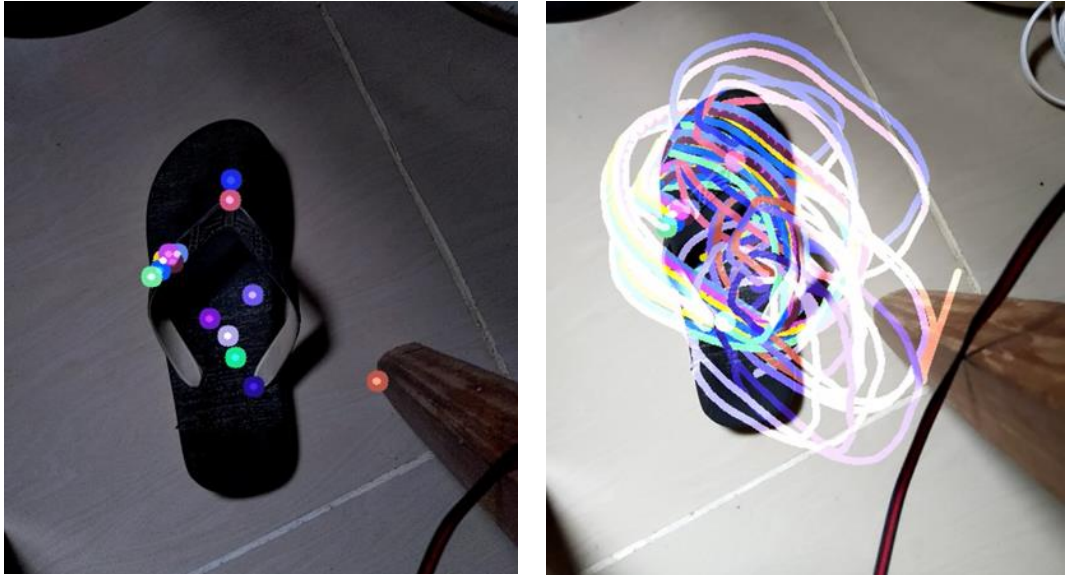

*Figure 2.* (a) Shi-Tomasi corner detection, left; (b) Lucas-Kanade optical flow, right.

After identifying the features to track, the Lucas-Kanade sparse optical flow algorithm was used to track the motion flow of the features found in the Shi-Tomasi method, Figure 2b (Lucas & Kanade, 1981; Choi et al., 2022). This technique compares two consecutive frames and assumes that they are separated by two small time increments (dt) such that objects between frames have moved a small distance from the origin. This also presumes that the intensity of chosen pixels does not change and does have similar motion (Hua et al., 2018). This motion tracking technique was used to estimate the position of the quadcopter during optical flow position hold.

### *Displacement calculation and position hold estimation*

Figure 3 shows the method used to obtain the x and y displacement of the quadcopter. This begins by solving the distance of the feature being tracked from the center of the frame using Euclidian distance (4) (Gandla et al., 2020; Wageeh et al., 2021). The x value when the tracked feature is on the left side is negative and positive when on the right side, computed using (2). Similarly, the y value is negative in the upper half of the image and positive in the bottom half, using (3).

$$x = x2 - x1 \hspace{4cm} \text{Equation 2}$$

$$y = y2 - y1 \qquad \text{Equation 3}$$

$$d = (x2 + y2)^{1/2} \qquad \text{Equation 4}$$

The x and y outputs are then fed into a PID controller algorithm to estimate the pitch and roll angle required to maintain the quadcopter position (Leal et al., 2021; Tagay et al., 2021). PID controllers are control loop mechanisms for applications needing continuous modulated control, as shown in Figure 4 ("PID controller," 2023). The PID controller can be represented mathematically by (5):

$$u(t) = K_\mathrm{p} e(t) + K_\mathrm{i} \int_0^t e(\tau)\,\mathrm{d}\tau + K_\mathrm{d} \frac{\mathrm{d}e(t)}{\mathrm{d}t}, \qquad \text{Equation 5}$$

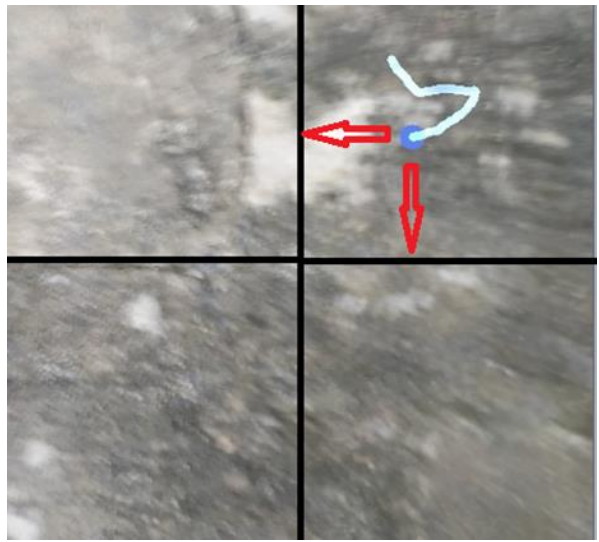

*Figure 3.* The current feature is tracked relative to the x and y center positions.

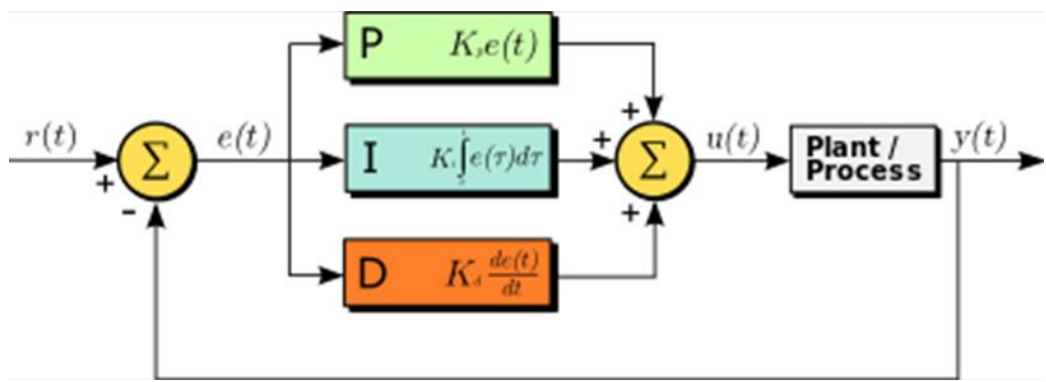

*Figure 4.* PID controller (Source: https://en.wikipedia.org/wiki/PID_controller)

## RESULTS AND DISCUSSION

A mobile app containing the optical flow function described in this paper was developed to test this concept. It was installed on the phone mounted on a quadcopter, as shown in Figure 5. The back camera of the phone is facing downward. The optical flow was automatically activated during take-off, together with the altitude hold. During tests, take-off was initiated using a remote application from a laptop computer connected to the phone via Wi-Fi.

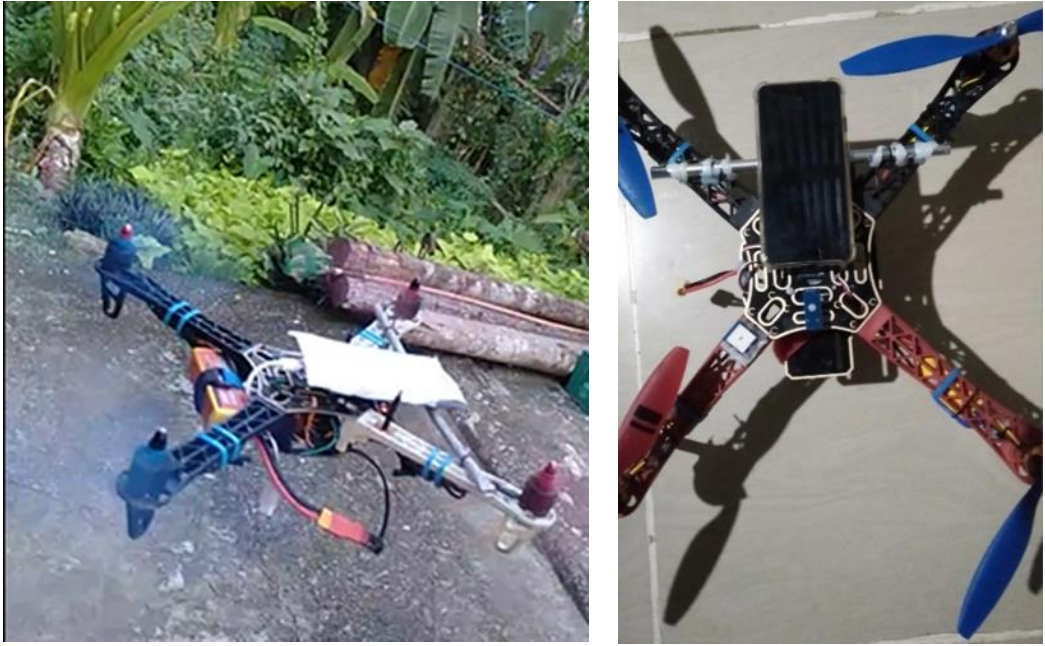

*Figure 5.* Quadcopters that were utilized during testing.

Actual flights were performed outdoors and indoors to test the quadcopter's optical flow position hold performance. In each test, the quadcopter hovered in the air at around 1 meter above the ground for five minutes. Its position was tracked during the entire flight, shown in Figure 6 - 7. The outdoor test shows a double standard deviation of 18.66 cm from the center. The quadcopter size is 58cm x 58cm (tip to tip of propeller). This implies that 95% of the time, the quadcopter is within a diameter of 96 cm during outdoor position hold. For indoor tests, a double standard deviation of 10.55 cm implies that 95% of the time, the quadcopter is within a diameter of 79 cm. The optical flow algorithm works as expected, allowing the quadcopter to perform pitch and roll maneuvers to maintain its position.

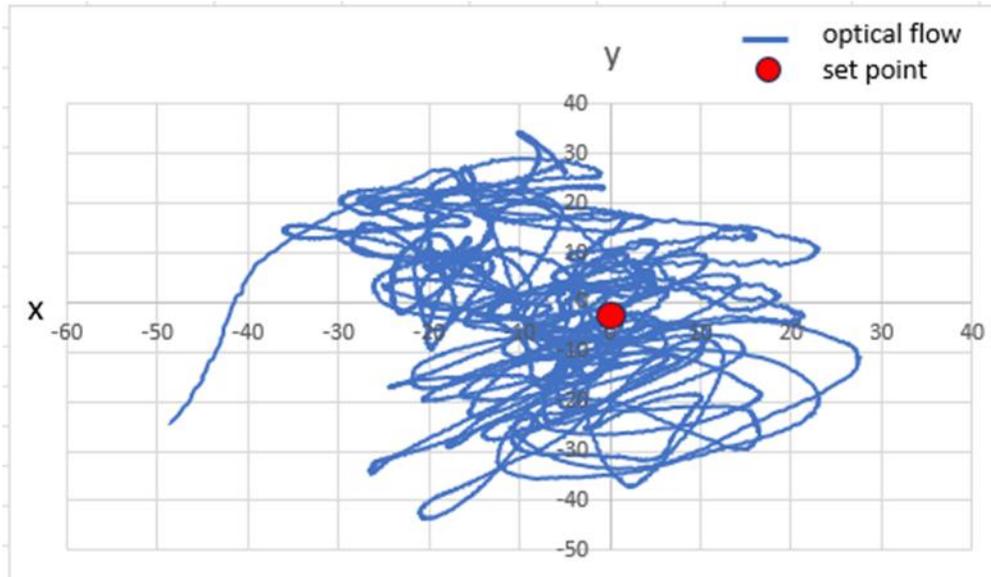

*Figure 6.* Optical flow position hold (outdoor)

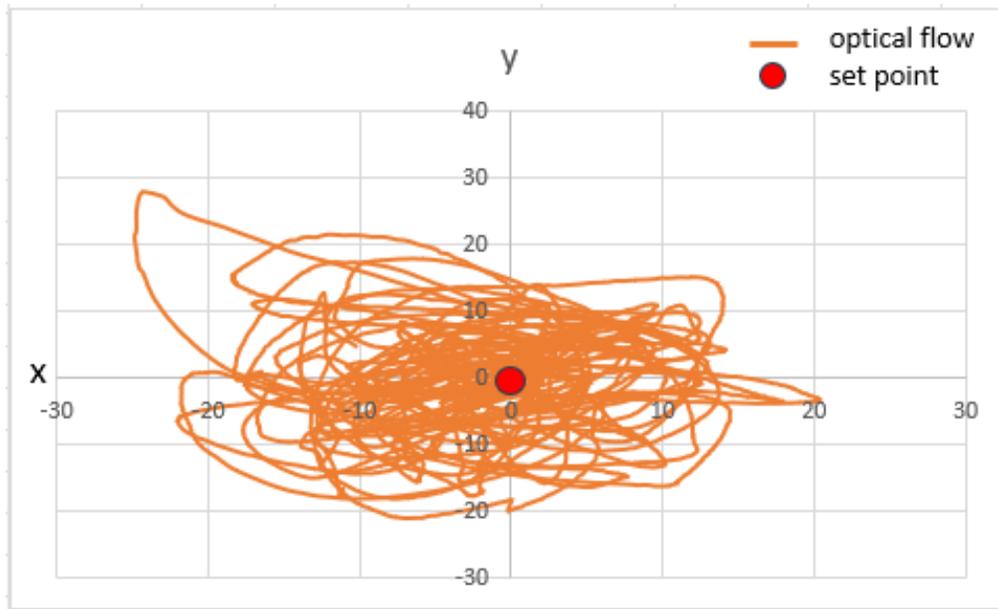

*Figure 7.* Optical flow position hold (indoor)

In this optical flow, multiple features are being tracked simultaneously, allowing the quadcopter to select the best feature to track and replace it when the algorithm cannot recognize it anymore due to drift, yaw rotation, or a change in intensity. During drift, where the previous features are no longer found, the algorithm automatically acquires another set of features to track, as shown in Figure 8.

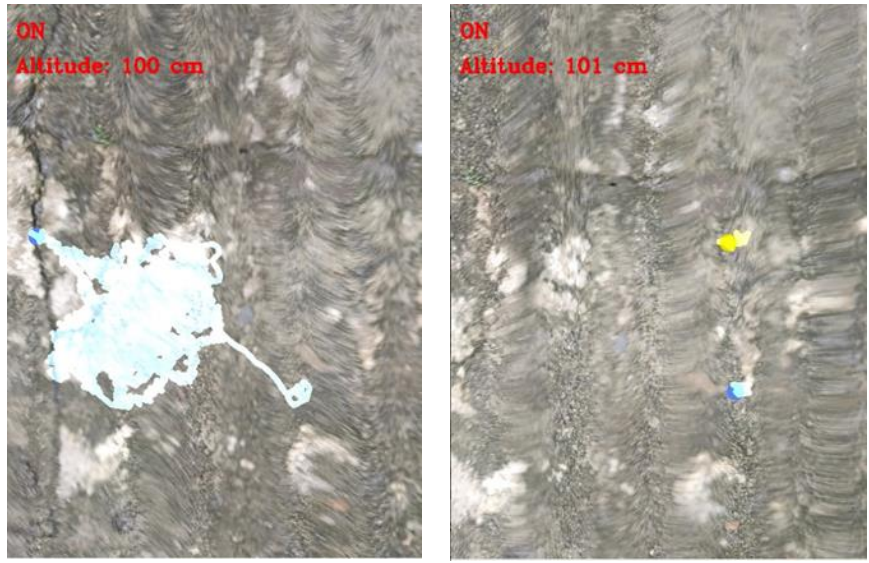

*Figure 8.* (a) Initial set of tracked features (left); (b) Acquired new set of tracked features after drift (right)

During low-light scenarios, the quadcopter performs satisfactorily as long as there are features that the camera can detect. In the total absence of visible features, the quadcopter is blind. Figure 9 shows how the quadcopter performs in low-light conditions.

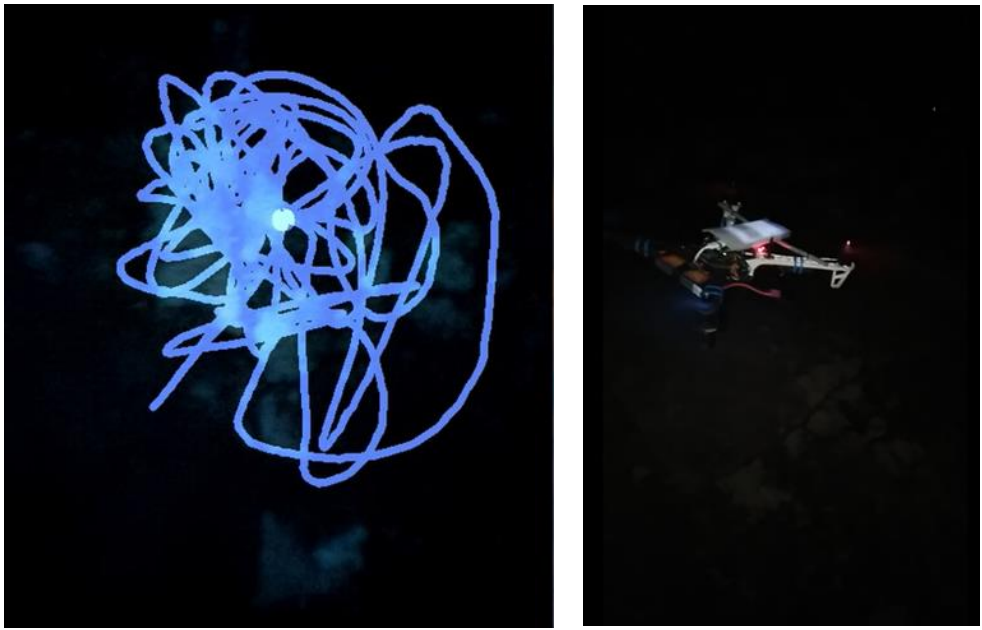

*Figure* 9. Optical flow in a low-light condition: (left) Camera view; (right) drone during position-hold flight

Most commercial drones advertise their vision hover accuracy as ±0.3 meters (see [www.dji.com](www.dji.com)), using dedicated vision sensors operating at over 100Hz. Using that as a performance baseline, the optical flow position-hold presented in this paper proves to be acceptable despite the limitation of the smartphone camera refresh rate of only 20-30Hz.

## CONCLUSIONS AND RECOMMENDATIONS

A smartphone camera's optical flow can maintain quadcopter position-hold in minimal natural disturbance. However, it tends to drift during strong winds, forcing it to acquire new features to track. Given the result of flight tests, compared with commercial drones, which are advertised to have a vision positioning hover accuracy of 0.3 meters, this smartphone-based position-hold performance is acceptable. Combining this with a GNSS source having absolute positioning information, sensor fusion can significantly improve the positioning algorithm of a phone-based quadcopter system.

## RESEARCH IMPLICATIONS

The drone functionality described in this paper performed well even though sensors found on smartphones are not explicitly designed to do this purpose. As different devices are integrated into mobile phones one by one, this paper presents an opportunity for phone manufacturers to tailor the device to be capable of driving a drone not as a remote control only but as its primary computing device, even at least improve the quality of sensors integrated into the device presents another potential smartphone use-case soon.

## ACKNOWLEDGEMENT

The researchers thank Iloilo State University of Fisheries Science and Technology, Cebu Institute of Technology University, and the Commission on Higher Education for making this undertaking possible.

## FUNDING

This study is funded by the Commission on Higher Education.

## DECLARATIONS

### *Conflict of Interest*

The researcher declares no conflict of interest in this study.

*Informed Consent*

This research has no respondents and, thus, contains no information that requires consent from participants.

*Ethics Approval*

This is not applicable since the execution of the research requires no respondents.

**Author's Biography**

Noel P. Caliston completed his Master in Information Technology at Cebu Institute of Technology University, Cebu City, Philippines, and is currently in the dissertation phase of the Doctor in Information Technology at the same university. He is also a faculty of the College of Information and Communication Technology at Iloilo State University of Fisheries Science and Technology – Dingle Campus, Dingle, Iloilo, Philippines. His teaching areas are computer programming and emerging technologies. His research interests are embedded systems, machine learning, and artificial intelligence.

Chris Jordan G. Aliac finished his bachelor's degree in Computer Engineering, master's in Computer Science, and doctorate in Information Technology at Cebu Institute of Technology University, Cebu City, Philippines. He is also a certified cloud practitioner by Amazon Web Service and a Certified Professional Computer Engineer by the Philippine Computer Engineering Certification Board. He focuses his research on artificial intelligence, specializing in Robotics and Machine Learning, Embedded and Distributed

Systems, and ICT Security. He is also the CIT University's Makespace manager and CIT University's ICT security Head. He is also a Full Professor at the College of Computer Studies at the same University.

James Arnold E. Nogra is an experienced industry practitioner in mobile, web, and full-stack development. He finished his bachelor's degree in Computer Science at the University of the Philippines. He completed his master's in Computer Science and doctorate in Information Technology at Cebu Institute of Technology University, Cebu City, Philippines. He also teaches subjects related to computer programming and data mining. His research interests are in the field of neural networks.